%% file: index.tex
\documentclass[11pt]{article}

\usepackage{multirow}
\usepackage{tabularx}
\usepackage{longtable}

\usepackage{listings}
\usepackage{color}
\usepackage{epsfig}
\usepackage{graphicx}
\usepackage[font=small]{subcaption} %% added
\usepackage[font=small]{caption}

\usepackage{amssymb}

\usepackage{amsmath}

\usepackage{hyperref}
 
\definecolor{codegreen}{rgb}{0,0.6,0}
\definecolor{codegray}{rgb}{0.5,0.5,0.5}
\definecolor{codepurple}{rgb}{0.58,0,0.82}
\definecolor{backcolour}{rgb}{0.95,0.95,0.92}

\lstdefinestyle{mystyle}{
    backgroundcolor=\color{backcolour},   
    commentstyle=\color{codegreen},
    keywordstyle=\color{magenta},
    numberstyle=\tiny\color{codegray},
    stringstyle=\color{codepurple},
    basicstyle=\footnotesize,
    breakatwhitespace=false,         
    breaklines=true,                 
    captionpos=b,                    
    keepspaces=true,                 
    numbers=left,                    
    numbersep=5pt,                  
    showspaces=false,                
    showstringspaces=false,
    showtabs=false,                  
    tabsize=2
}
 
\usepackage[affil-it]{authblk} 
\usepackage{etoolbox}
\usepackage{lmodern}

\makeatletter
\patchcmd{\@maketitle}{\LARGE \@title}{\fontsize{16}{19.2}\selectfont\@title}{}{}
\makeatother

\usepackage{authblk}
\author[1,2]{Dario Zanca\thanks{Corresponding author: dario.zanca@unifi.it}}
\affil[1]{Department of Information Engineering, University of Florence, Florence, Italy}
\affil[2]{Department of Information Engineering and Mathematics, University of Siena, Siena, Italy}
\author[3]{Valeria Serchi}
\author[3]{Pietro Piu}
\author[3,4]{\\Francesca Rosini}
\author[3,4]{Alessandra Rufa}
\affil[3]{Eye-tracking and Visual Application Lab (EVALab), Department of Medicine, Surgery and Neurosciences, University of Siena, Siena, Italy}
\affil[4]{Neurological and Neurometabolic Unit, Department of Medicine, Surgery and Neurosciences, University of Siena, Siena, Italy}
\date{}                     %% if you don't need date to appear

\begin{document}

% \author{Dario Zanca, ***}
\title{FixaTons: A collection of Human Fixations\\ Datasets and Metrics for Scanpath Similarity}
\maketitle

    \newpage

    \tableofcontents
    
    \newpage

    \section{Introduction}
    In the last three decades, human visual attention has been a topic of great interest in various disciplines. In computer vision, many models have been proposed to predict the distribution of human fixations on a visual stimulus. Recently, thanks to the creation of large collections of data, machine learning algorithms have obtained state-of-the-art performance on the task of saliency map estimation.
    
    On the other hand, computational models of scanpath are much less studied. Works are often only descriptive \cite{leestella} or task specific \cite{renniger}. This is due to the fact that the scanpath is harder to model because it must include the description of a dynamic.  General purpose computational models are present in the literature, but are then evaluated in tasks of saliency prediction \cite{zanca},  losing therefore information about the dynamics and the behaviour. In addition, two technical reasons have limited the research. The first reason is the lack of robust and uniformly used set of metrics to compare the similarity between scanpath. The second reason is the lack of sufficiently large and varied scanpath datasets.
    
    In this report we want to help in both directions. We present \textit{FixaTons}, a large collection of datasets human scanpaths (temporally ordered sequences of fixations) and saliency maps. It comes along with a software library for easy data usage, statistics calculation and implementation of metrics for scanpath and saliency prediction evaluation.

    \input{datasets}

\input{online.tex}

\input{structure.tex}

    \input{software.tex}    
    
    \input{metrics.tex}

\input{BIB.tex}
\end{document}

%% file: datasets.tex
\section{Datasets included in the collection}
FixaTons is a collection of publicly available datasets of human fixations. The authors of this paper have also included SIENA12, which is a further set of grayscale images that aim at being as more free of semantic content as possible.

    \subsection{SIENA12}
    The SIENA12 dataset includes 12 grayscale images\footnote{The authors thank Danilo Pileri for kindly providing images of the dataset.}. It was collected by the author of this paper in collaboration with EVALAB at Policlinico Alle Scotte in Siena. The images have been selected so as to minimize the semantic content of the scenes. Images include natural scenes, human constructions, but also abstract contents.
    
    \begin{table}[h]
        \centering
        \begin{tabular}{|l|l|}
            \hline
             Dataset Name & SIENA12 \\
            \hline
             Number of images & 12 \\
            \hline
             Size & 1024x768 px \\
            \hline
             Categories & Outdoor, natural, synthetic \\
            \hline
             Number of observers & 23 \\
            \hline
             Age of the observers & From 23 to 52 \\
            \hline
             Task & Free-viewing \\
            \hline
             Duration & 5 seconds \\
            \hline
             Eye-tracker & ASL 504 (240 Hz) \\
            \hline
             Screen & LCD 1024 $\times$ 768 px (31 $\times$ 51 cm) \\
            \hline
             Eye-screen distance & 72 cm \\
            \hline
             Other information & Grayscale images \\
            \hline
        \end{tabular}
        \caption{Tech. spec. of the dataset SIENA12}
        \label{tab:my_label}
    \end{table}

\begin{figure}[h]
	\begin{center}
			\includegraphics[width=.88\linewidth]{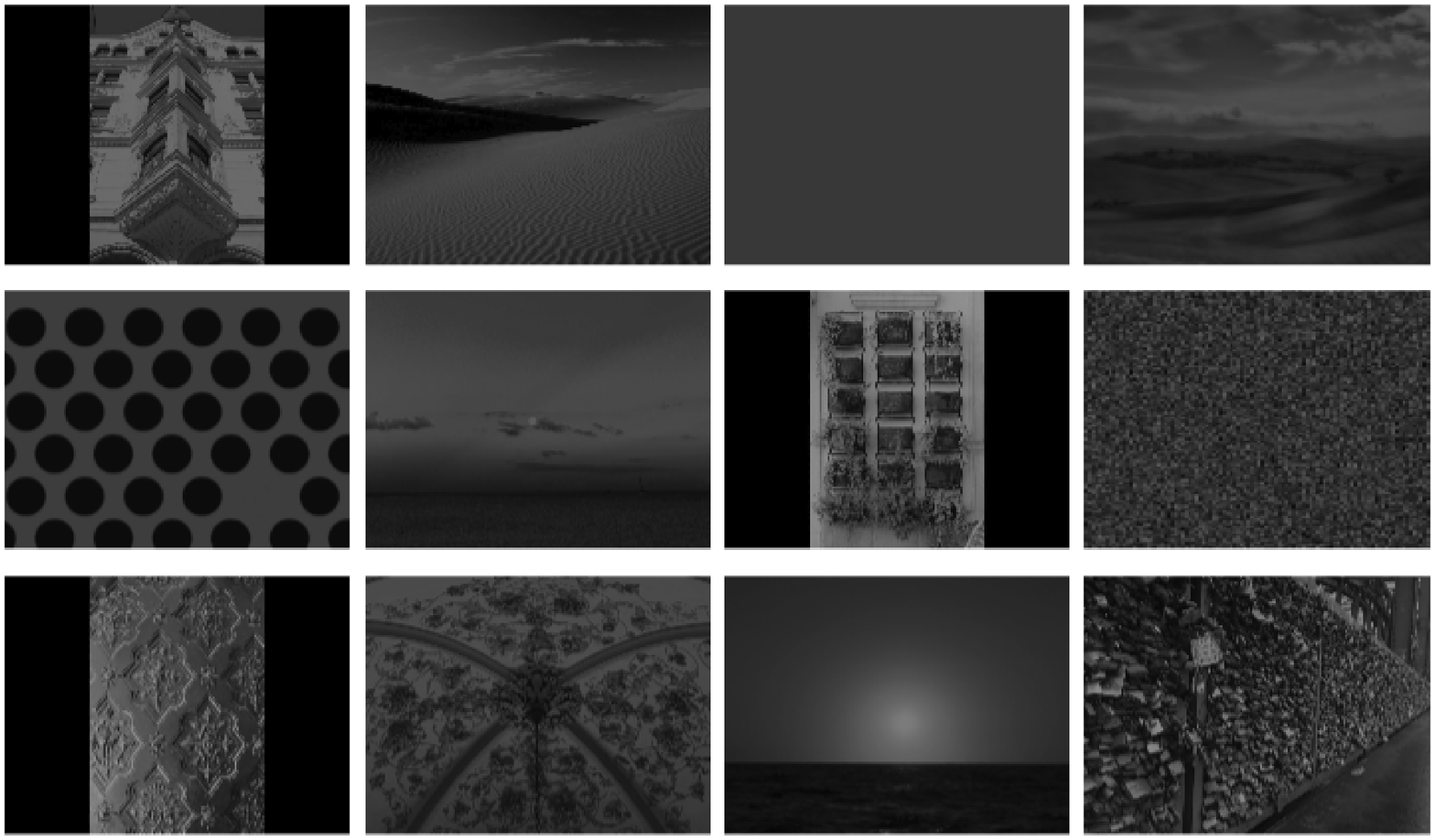}
	\end{center}
	\vspace{-1.5em}
	\caption{\textbf{Images from SIENA12} }
	\label{fig:siena12}
\end{figure}

    \subsubsection{Protocol for data collection in SIENA12}
    Visual data are collected through a $240$ Hz eye-tracking system (ASL 504, Applied Science Laboratories, Bedford, MA, USA) allowing for a remote tracking of the point of gaze on a calibrated surface (LCD screen, $1024 \times 768$ px, $31 \times 51$ cm). A chin-rest is used to maintain constant the relative distances between the eyes of the subject, the eye-tracker optics and the screen (eye/eye-trackers distance, $68$ cm; eye/screen distance, $72$ cm). The height of the chin-rest is set in order to get the line of sight of the subjects at the rest position perpendicular to the centre of the screen. The stimuli presentation and the data collection is managed by customized software. All recordings are conducted in complete darkness, measuring one eye.\\
    A nine-point calibration is performed trough an interactive user interface provided by the manufacturer. The operator instructs the participant to perform 5 seconds tasks of free visual exploration. A total of 12 images is presented. Between one image and another, a central dot (of about 2 seconds) is displayed. The order of the images composing the sequence of the experimental stimuli is randomly chosen to prevent bias.\\
    Raw data has been processed with the publicly available python library PyGazeAnalyser~\cite{pygazeanalyzer} in order to extract information about fixations. 
    
    \newpage 
    \subsection{Other Datasets}
    Authors of compatible datasets are encouraged to add them to this collection. Please, send an email to the corresponding author of this report along with an authorization to redistribute that data. Below, the list of datasets currently added to the collection.
    
    \input{dataset_mit1003.tex}

%% file: dataset_mit1003.tex
    \subsubsection{MIT1003}
    The dataset MIT1003~\cite{mit1003} is a large database of eye tracking data. The images, eye tracking data, and accompanying code in Matlab are all available on the web. This dataset can be used as training data for the MIT300~\cite{mit300} benchmark since they share the same technical specification. MIT300 is not included in FixaTons collection because its data is kept private for a fair evaluation of the benchmark. More details about the protocol used for data collection are given in the referred paper.
        \newpage

    \begin{table}[h]
        \centering
        \begin{tabular}{|l|l|}
            \hline
             Dataset Name & MIT1003 \\
            \hline
             Number of images & 1003 \\
            \hline
             Size & Min. dim.: 405 px -- Max. dim.: 1024 px\\
            \hline
             Categories & Landscape and portrait images \\
            \hline
             Number of observers & 15 \\
            \hline
             Age of the observers & From 18 to 35 \\
            \hline
             Task & Free-viewing \\
            \hline
             Duration & 3 seconds \\
            \hline
             Eye-tracker & ETL 400 ISCAN (240Hz) \\
            \hline
             Screen & LCD 1024 $\times$ 768 px (40.5 $\times$ 30 cm) \\
            \hline
             Eye-screen distance & 75 cm \\
            \hline
             Other information & It can be used as training data for MIT300 benchmark \\
            \hline
        \end{tabular}
        \caption{Tech. spec. of the dataset MIT1003}
        \label{tab:mit1003description}
    \end{table}

%% file: online.tex
    \section{Online resources}
    \begin{itemize}
        \item Webpage of the project: \url{http://sailab.diism.unisi.it/fixatons/}
        \item Data download: \url{https://drive.google.com/open?id=1TQSaq5J0p_oCdkyVZ-IzBltLwJ2cm3UA}
        \item Software library \url{https://github.com/dariozanca/FixaTons}.
    \end{itemize}
    \newpage

%% file: structure.tex
    \section{Structure of the FixaTons collection}
    
    \begin{itemize}
        \item FixaTons
        \begin{itemize}
            \item DATASET\_NAME
            \begin{itemize}
                \item STIMULI : contains original images.
                  They can have different file format (jpg, jpeg, png,...)

                \item SCANPATHS : contains one folder for each image

                \begin{itemize}
                    \item IMAGE\_ID :
                  it contains one file for each scanpath of that image
                  scanpaths are matrices
                  rows of this matrices describe fixations
                  each fixation is of the form :
                  [x-pixel, y-pixel, initial time, final time].
                  Times are in seconds.
                \end{itemize}

                \item FIXATION\_MAPS : contains a fixation map of each original image
            they are matrices of zeros (non-fixated pixels) and ones (fixated
            pixels). They can have different file format (jpg, jpeg, png,...)

                \item SALIENCY\_MAPS : contains saliency maps of each original image             they are generated from human data. They can have different file
            format (jpg, jpeg, png,...) 
            \end{itemize}
        \end{itemize}
    \end{itemize}

%% file: software.tex
\section{Software included}
Some software tools are provided together with the collection for an easy use and visualization of the data. 

Software is written in python. All the functions are included in the file \textit{FixaTons.py}. They make use of the public library OpenCV which should be installed on the machine before the use of \textit{FixaTons.py}.

Functions can be divided in five main categories:
\begin{itemize}
    \item List information
    \item Get data (matrices)
    \item Visualize data
    \item Compute metrics.
    \item Compute statistics
\end{itemize}

    \subsection{List information}
    The collection comprehend different datasets, each of them with different stimuli names, number of subjects,, subjects id's, etc. The provided software allows to easily get this kind of information.
    
    \begin{itemize}
        \item \textbf{FixaTons.list.dataset()}: This functions returns a list with the names of the datasets included in the collection.
    
        \item \textbf{FixaTons.list.stimuli(DATASET\_NAME)}: This functions lists the names of the stimuli of a specified dataset. 
        
        \item \textbf{FixaTons.list.subjects(DATASET\_NAME, STIMULUS\_NAME)}: This functions lists the id's of the subjects which have been watching a specified stimuli of a dataset.
    \end{itemize}

    \subsection{Get data (matrices)}
    
    Different functions allows to get data in form of numpy array.
    
    \begin{itemize}
    
        \item \textbf{FixaTons.get.stimulus(DATASET\_NAME, STIMULUS\_NAME)}: This functions returns the matrix of pixels of a specified stimulus. Notice that, both \\DATASET\_NAME and STIMULUS\_NAME need to be specified. The latter, must include file extension. The returned matrix could be 2- or 3-dimensional. 
    
        \item \textbf{FixaTons.get.fixation\_map(DATASET\_NAME, STIMULUS\_NAME)}: This functions returns the matrix of pixels of the fixation map of a specified stimulus. Notice that, both DATASET\_NAME and STIMULUS\_NAME need to be specified. The latter, must include file extension. The returned matrix is a 2-dimensional matrix with 1 on fixated locations and 0 elsewhere. 
    
        \item \textbf{FixaTons.get.saliency\_map(DATASET\_NAME, STIMULUS\_NAME)}: This functions returns the matrix of pixels of the saliency map of a specified stimulus. Saliency map has been obtained by convolving the fixation map with a proper gaussian filter (corresponding to one degree of visual angle). Notice that, both \\DATASET\_NAME and STIMULUS\_NAME need to be specified. The latter, must include file extension. The returned matrix is a 2-dimensional matrix.
    
        \item \textbf{FixaTons.get.scanpath(DATASET\_NAME, STIMULUS\_NAME, subject = 0)}: This functions returns the matrix of fixations of a specified stimulus. The scanpath matrix contains a row for each fixation. Each row is of the type \textit{[x, y, initial\_t, final\_time]}. By default, one random scanpath is chosen between available subjects. For a specific subject, it is possible to specify its id on the additional argument subject=id. 
    \end{itemize}
    
    \subsection{Visualize data}
    
    For an easy visualization of the data, some functions have been included in the library.
    
    \begin{itemize}
        
        \item \textbf{FixaTons.show.map(DATASET\_NAME, STIMULUS\_NAME, showSalMap = True, showFixMap = False, plotMaxDim = 0)}: This functions uses cv2 standard library to visualize a specified stimulus. By default, stimulus is shown with its saliency map aside. It is possible to deactivate such option by setting the additional argument showSalMap=False. It is possible to show also (or alternatively) the fixation map by setting the additional argument showFixMap=True. Depending on the monitor or the image dimensions, it could be convenient to resize the images before to plot them. In such a case, user could indicate in the additional argument plotMaxDim=500 to set, for example, the maximum dimension to 500. By default, images are not resized.
    
        \item \textbf{FixaTons.show.scanpath(DATASET\_NAME, STIMULUS\_NAME, subject = 0, animation = False, putNumbers = True, plotMaxDim = 0)}: This functions uses cv2 standard library to visualize the scanpath of a specified stimulus. By default, one random scanpath is chosen between available subjects. For a specific subject, it is possible to specify its id on the additional argument subject=id. It is possible to visualize it as an animation by setting the additional argument animation=True. Depending on the monitor or the image dimensions, it could be convenient to resize the images before to plot them. In such a case, user could indicate in the additional argument plotMaxDim=500 to set, for example, the maximum dimension to 500. By default, images are not resized.
            
    \end{itemize}
    
\subsection{Compute metrics}
    An implementation of the most common metrics to compute saliency maps similarity and scanpaths similarity is included in the software provided with FixaTons. A mathematical description of the metrics of scanpath similarity is postponed to the Appendix.
    \begin{itemize}
        \item Saliency Map similarities
            \begin{itemize}
                \item \textbf{FixaTons.metrics.KLdiv(saliencyMap1, saliencyMap2)}: This function computes the Kullback–Leibler divergence between two continuous saliency maps.
                \item \textbf{FixaTons.metrics.AUC\_Judd(saliencyMap, fixationMap, jitter = True, toPlot = False)}
                Given a continuous saliency map (normally the output of a saliency model) and a fixation map (matrix with 1's at fixated locations, 0's elsewhere), it computes the Area Under the ROC curve, in the implementation described by Judd in \cite{mit1003}.
                \item \textbf{FixaTons.metrics.NSS(saliencyMap, fixationMap)}
                Given a continuous saliency map (normally the output of a saliency model) and a fixation map (matrix with 1's at fixated locations, 0's elsewhere), it computes the Normalized Scanpath Saliency
            \end{itemize}
        \item Scanpaths similarities
            \begin{itemize}
                \item \textbf{FixaTons.metrics.euclidean\_distance(human\_scanpath, simulated\_scanpath)}: This function computes the euclidean distance between two scanpaths. More details are given on the Appendix.
                \item \textbf{FixaTons.metrics.string\_edit\_distance(stimulus, human\_scanpath, simulated\_scanpath, n = 5, substitution\_cost=1)}: This function computes the string-edit distance between two scanpaths. More details are given on the Appendix.
                \item \textbf{FixaTons.metrics.time\_delay\_embedding\_distance( human\_scanpath, simulated\_scanpath, k = 3, distance\_mode = 'Mean')}: This function computes the time-delay embedding distance between two scanpaths. More details are given on the Appendix.
                \item \textbf{FixaTons.metrics.scaled\_time\_delay\_embedding\_distance( human\_scanpath, simulated\_scanpath, image, toPlot = False))}: This function computes the scaled time-delay embedding distance between two scanpaths. More details are given on the Appendix.
            \end{itemize}
    \end{itemize}
    
\subsection{Compute statistics}
    
    It is possible to compute statistics for the overall collection, or for a specific dataset, about some scanpath properties.
    
    \begin{itemize}
        \item \textbf{FixaTons.stats.statistics(DATASET\_NAME=None)}: This functions returns a list with two values: fixations per second and the average of saccades length. If no dataset is specified, statistics are calculated on the whole FixaTons collection. To restrict computation on a specific dataset, it is sufficient to indicate its name on the additional argument DATASET\_NAME. 
    \end{itemize}
    
\subsection{Example of use}
    Here we propose a python script which show a complete example of use of some facilities.
    \begin{lstlisting}[language=Python, caption=Example of use. Complete Python script.]
# import the library
import FixaTons

#shuffle(dataset_list)
for dataset in FixaTons.list.datasets():

    # For all images in that dataset
    for image in FixaTons.list.stimuli(dataset):

        # Show the image aside its saliency map (5 seconds dy default).
        FixaTons.show.map(dataset, image, plotMaxDim=1500)

        # Then, for all the subjects that watched that image,
        for subject in FixaTons.list.subjects(dataset, image):

            # Show the correspondent scanpath as an animation.
            # (Look, time of exploration in the animation is the
            # exact time, from the dataset.)
            FixaTons.show.scanpath(dataset, image, subject,
                                   animation=True,
                                   plotMaxDim=1000,
                                   wait_time=1000)
                                   
\end{lstlisting}

%% file: metrics.tex
    \section{Appendix A: }

    In this section we define metrics to compute similarity between scanpaths. Metrics of course work for given two sequences of point in the $\mathbb{R}^2$ plane, but for the sake of clearness, we will contextualize to our topic. Let say our goal is to compute a distance between a \textit{simulated} and a \textit{human} scanpath,
	    
	    $$s \equiv s(1), ..., s(n)$$
	    $$h \equiv h(1), ..., h(m)$$
	
	\noindent where 
	
		$$s(i) \in \mathbb{R}^2, \forall i \in \{1,...,n\}, $$
		$$h(j) \in \mathbb{R}^2, \forall j \in \{1,...,m\}.  $$

    \subsection {Euclidean distance}
    
    Let us suppose the two scanpaths $s$ and $h$ have the same legth $n$. We define the \textit{Euclidean distance} between two scanpath as the mean of the Euclidean distances between corresponding fixations. To give a mathematical formulation, 
    
    \begin{equation}\label{euclidean}
        d_{Euclidean}(s,h) = \frac{1}{n}\sum_{i = 0}^n \Vert s(i) - h(i) \Vert_2
    \end{equation}
    
	\noindent This metric presents some important drawbacks.
    
    \begin{itemize}
        \item[i] The two scanpaths has to have the same length, \textit{i.e.} $n$ must be equal to $m$. But human scanpaths in the available collection of data rarely have same length, and to restrict to the minimum length would produce a huge loss of important data.
        \item[ii] Euclidean metric does not take time-shifts into account. Let us indicate fixation locations $s(i), h(i) \in \mathbb{R}^2, \forall i \in \{1, ..., n\}$, with capital letters. Then, the scanpaths ABCDEF and ZABCDE, even if very similar, would be very distant.
    \end{itemize}

    \subsection{String-edit similarity}
    
   One of the most common metrics to compute scanpaths similarity is the "string-edit" or Levenshtein distance. The algorithm of the dynamic program to compute this metric is described in~\cite{Jurafsky}. The metric has been used in different works for comparing different human scanpaths~\cite{Foulsham, Brandt}. It has been shown~\cite{Choi} to be robust to changes in the number of regions used.
   
   Input stimulus is divided into $n x n$ regions, labeled with characters. Scanpaths are turned into strings by associating the corresponding character to each fixation. Finally, the algorithm of string-edit is used to measure the distance between the two generated strings.

    \subsection {Time-delay embedding distance} 
    Time-delay embedding are used in order to quantitatively compare stochastic and dynamic scanpaths of varied lengths. This metric is largely used in dynamic systems analysis and carefully described in~\cite{wang2011simulating}. In particular, let us consider the problem of comparing a \textit{simulated} and a \textit{human} scanpath,
    
	    $$s \equiv s(1), ..., s(n)$$
	    $$h \equiv h(1), ..., h(m)$$
	    
   	\noindent eventually of different length, $n \ne m$. They are ordered list of fixations. Each fixation is defined by a couple of spatial coordinates (that correspond to the fixated location in the image), that is
    
		$$s(i) \in \mathbb{R}^2, \forall i \in \{1,...,n\}, $$
		$$h(j) \in \mathbb{R}^2, \forall j \in \{1,...,m\}.  $$
	
	\noindent We indicate with 
	
		$$\mathcal{C}_s^k(t) = (s(t), ..., s(t+k))$$ 
	
	\noindent a $k$-dimensional sub-sequence of fixation extracted from the simulated scanpath $s$, starting from the $t$-th fixation, and with
	
		\begin{equation}
		X = \{ \mathcal{C}_s^k(t) \}_{t \in (1, ..., n-k)} \subset \mathbb{R}^k
		\end{equation}	
		
	\noindent the space of all this possible $k$-dimensional sub-sequences extracted from $s$. Analogously,
	
	$$\mathcal{C}_h^k(t) = (h(t), ..., h(t+k))$$ 
	
	\noindent is a $k$-dimensional sub-sequence of fixation extracted from the human scanpath $h$, starting from the $t$-th fixation, and 
		
	\begin{equation}
	Y = \{ \mathcal{C}_h^k(t) \}_{t \in (1, ..., m-k)} \subset \mathbb{R}^k
	\end{equation}
	
	\noindent is the the space of all this possible $k$-dimensional sub-sequences extracted from $h$. Notice that $k < \min\{n,m\}$. Comparison between the clouds $X$ and $Y$ of data points in $\mathbb{R}^k$ will reflect dynamical similarities between the two scanpaths.
	
	\noindent Given 
	
	\begin{equation}\label{tde_base_distance}
	d_k(x, Y) = \min_{y \in Y}{ \{ \Vert x - y \Vert \} },
	\end{equation}
	
	\noindent the time-delay embedding distance is then defined as the \textit{mean minimal} distance (MM-distance)
	
		\begin{equation}\label{MM_distance}
			tde_k^{MM}(s, h) = \frac{1}{|X|} \sum_{x \in X} d_k(x, Y)
		\end{equation}
		
	\noindent or as the Hausdorff distance (H-distance)
		
		\begin{equation}\label{H_distance}
			tde_k^{H}(s, h) = \max_{x \in X} d_k(x, Y).
		\end{equation}

	\subsection{Scaled Time-delay embedding distance}
	
	The distance based on the time-delay embedding described in the previous section allows to capture dynamics as well as stochastics aspects of the compared processes. The resulting metric is, however, given in pixel, since no re-scaling is applied anywhere. In real world datasets input may differ a lot in resolution. To take this into account, we define a scaled version of such a metric, where fixations coordinates are normalized in $\left[0, 1\right]^2$.
	
	More specifically, let us consider the problem of comparing a \textit{simulated} and a \textit{human} scanpath over an image $I$, say
	
		$$s \equiv s(1), ..., s(n)$$
		$$h \equiv h(1), ..., h(m)$$
		
	\noindent eventually of different length, $n \ne m$. They are ordered list of fixations. Each fixation is defined by a couple of spatial coordinates, that is
		
		$$s(i) \in \left[0, l_1\right], \forall i \in \{1,...,n\}, $$
		$$h(j) \in \left[0, l_2\right], \forall j \in \{1,...,m\}. $$
		
	\noindent where $l_1$ and $l_2$ are, respectively, the horizontal and vertical dimensions of $I$. First, fixations coordinates are normalized in $\left[0, 1\right]^2$, dividing by maximum of the dimensions $L = max\{l_1, l_2\}$, that is
	  
		$$\bar{s}(i) = s(i)/L, \forall i \in \{1,...,n\}, $$
		$$\bar{h}(j) = h(i)/L, \forall j \in \{1,...,m\}. $$
		
	\noindent Now we notice that \textit{tde} depends on the choice of the sub-sequences lengths $k$. To obtain a single more comprehensive measure, we define the \textit{scaled time-delay embedding similarity} as
		
		\begin{equation}\label{stde}
			stde(s, h) = \exp^{-\frac{1}{|K|} \sum_{k \in K} tde_k(s, h)},
		\end{equation}
	
	\noindent where $K = \{ 1, ..., \min{ \{n,m\} } - 1 \}$ is the set of all possible sub-sequences lengths. Notice that $stde(s, h) = 1$ indicates perfect similarity, while as it approaches zero the more dissimilar the scanpaths are.